\documentclass[10pt,twocolumn,letterpaper]{article}

\usepackage{cvpr}              
\usepackage{amsmath,amssymb,amsfonts}
\usepackage{graphicx}
\usepackage{textcomp}
\usepackage{algpseudocode}
\usepackage{algorithm}
\usepackage[dvipsnames]{xcolor}

\definecolor{cvprblue}{rgb}{0.21,0.49,0.74}
\usepackage[pagebackref,breaklinks,colorlinks,citecolor=cvprblue]{hyperref}

\def\paperID{9} 
\def\confName{CVPR}
\def\confYear{2024}

\title{ALINA: Advanced Line Identification and Notation Algorithm}

\author{Mohammed Abdul Hafeez Khan\\
Florida Institute of Technology\\
Melbourne, Florida\\
{\tt\small mkhan@my.fit.edu}
\and
Parth Ganeriwala\\
Florida Institute of Technology\\
Melbourne, Florida\\
{\tt\small pganeriwala2022@my.fit.edu}
\and
Siddhartha Bhattacharyya\\
Florida Institute of Technology\\
Melbourne, Florida\\
{\tt\small sbhattacharyya@fit.edu}
\and 
Natasha Neogi\\
NASA Langley Research Center\\
Hampton, Virginia\\
{\tt\small  natasha.a.neogi@nasa.gov}
\and
Raja Muthalagu\\
BITS Pilani Dubai Campus\\
Dubai, UAE\\
{\tt\small raja.m@dubai.bits-pilani.ac.in}
}
\begin{document}
\twocolumn[{%
\renewcommand\twocolumn[1][]{#1}%
\maketitle
\begin{center}
    \centering \includegraphics[width=0.8\linewidth]{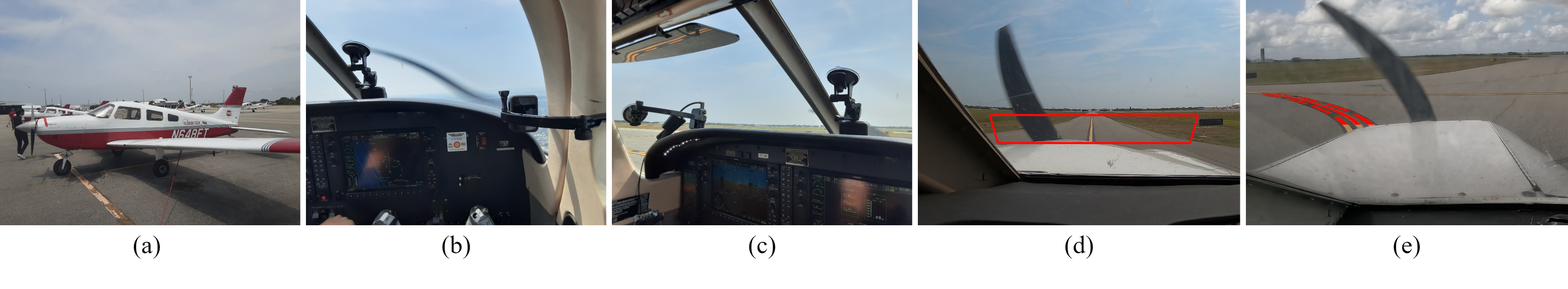}
    \captionof{figure}{(a) Illustrates the aircraft used for this research, (b) and (c) displays the setup of cameras at different perspectives inside the aircraft, (d) shows the manually defined region of interest on a frame, (e) showcases the annotation label created by ALINA on a frame.}
    \label{fig:main}
\end{center}%
}]
\begin{abstract}
Labels are the cornerstone of supervised machine learning algorithms. Most visual recognition methods are fully supervised, using bounding boxes or pixel-wise segmentations for object localization. Traditional labeling methods, such as crowd-sourcing, are prohibitive due to cost, data privacy, amount of time, and potential errors on large datasets. To address these issues, we propose a novel annotation framework, Advanced Line Identification and Notation Algorithm (ALINA), which can be used for labeling taxiway datasets that consist of different camera perspectives and variable weather attributes (sunny and cloudy). Additionally, the CIRCular threshoLd pixEl Discovery And Traversal (CIRCLEDAT) algorithm has been proposed, which is an integral step in determining the pixels corresponding to taxiway line markings. Once the pixels are identified, ALINA generates corresponding pixel coordinate annotations on the frame. Using this approach, 60,249 frames from the taxiway dataset, AssistTaxi have been labeled. To evaluate the performance, a context-based edge map (CBEM) set was generated manually based on edge features and connectivity. The detection rate after testing the annotated labels with the CBEM set was recorded as 98.45\%, attesting its dependability and effectiveness. Code: \url{https://github.com/hafeezkhan909/ALINA}. 
\end{abstract}
    
\section{Introduction}
\label{sec:intro}

With the advancement of sensors, hardware and intelligent software,  autonomous vehicles have become the center of attention for computer vision and robotics research. Among the several components of autonomous vehicles, the camera-based lane detection plays a crucial role in perceiving the lines and then assisting autonomous vehicular systems in making decisions to follow the correct path/lanes \cite{Cultrera_2020_CVPR_Workshops}. 

It is envisioned, as we gain confidence in the correctness of the design of autonomous systems, it will pave the way for the integration of learning based technologies in assisting safety critical systems such as, autonomous aircraft for urban air mobility or aerial delivery \cite{Bhattacharyya_2021,10131227}. As a result, this has increased the need for an advanced perception system, equipped with precise line identification capability. In future, it can assist pilots in the aircraft, especially when navigating the airport taxiway. As stated by the Aviation Safety Network \cite{Accidentstatistics}, approximately, 33.33\% of the aircraft accidents, between the years 2015 and 2022, happened during the taxiing phase. The common causes of taxiway accidents include poor weather conditions and high traffic on the taxiway. In order to address these issues, some researchers have formulated solutions for assisting pilots during the taxiing phase, including: light-based guidance systems, which have been designed by the Federal Aviation Administration and Honeywell \cite{airporttech,Stern_2001}, computer vision based taxiway guidance techniques \cite{Batra2020}, Airport Moving Maps \cite{Jeppesen_2021}, and sensors, such as LIDAR and cameras mounted on an aircraft \cite{Aircraft_Navigation_on_Taxiways,theuma2019}.       
This research builds on the work of Ganeriwala et. al \cite{ganeriwala2023assisttaxi}, particularly with the contour-based detection and line extraction method (CDLEM). They introduced AssistTaxi dataset, which incorporates more than 300,000 frames of taxiway and runway instances. The data was collected from Melbourne (MLB) and Grant-Valkaria (X59) general aviation airports. 

In this research, we introduce an Advanced Line Identification and Notation Algorithm (ALINA), which is an annotation framework developed to detect and label taxiway line markings from video frames (Fig. \ref{fig:main}). ALINA establishes a uniform trapezoidal region of interest (ROI) by utilizing the initial frame, that is consistently applied across all subsequent frames of the video. The ROI is then geometrically modified and the color space is transformed to produce a binary pixel map. ALINA pinpoints pixels representing taxiway markings through the novel CIRCular threshoLd pixEl Discovery And Traversal (CIRCLEDAT) algorithm, leading to frame annotations and coordinate data files. A context-based edge map (CBEM) set was generated for comparison to ensure accuracy in marking detection. ALINA was tested on a subset of AssistTaxi dataset - 60,249 frames extracted from three distinct videos with unique camera angles. The focus of this research had been on 60,249 frames out of a vast 300,000-frames AssistTaxi dataset, as the motivation was to lay down a rigorous yet tractable foundation for the empirical validation of ALINA's efficacy, paving the way for its scalability to larger datasets in future. Through this research, we primarily aim to reduce the intensive, expensive and error-prone manual labeling \cite{fredriksson2020data}, and provide labeled data to enhance taxiway navigation safety.

\textbf{Our contribution} include the following four main aspects:
\begin{itemize}
    \item To reduce the intensive, expensive, and error-prone manual labeling process, thus contributing to the efficient combination of automated and manual creation of labelled datasets.
    \item Development of the Advanced Line Identification and Notation Algorithm (ALINA) providing a robust framework for precise detection and labelling of continuous video datasets, particularly focusing on taxiway line markings.
    \item Introduction of a novel algorithm, CIRCLEDAT, tailored to pinpoint pixels representing taxiway/road markings with high accuracy, ensuring precise labeling across continuous video frames.
    \item Establishing a systematic approach based on change in perspective of scenario to justify the sample size of ground truth which is a subset from the AssistTaxi dataset \cite{ganeriwala2023assisttaxi}.
\end{itemize}
The remainder of the paper is organized as follows: 
In \cref{related}, related research works have been discussed. \cref{method} entails the detailed methodology of ALINA, \cref{results} presents the experimental results. Finally, \cref{conclude} provides the conclusion and outlines the directions for future work.

\section{Literature Review}
\label{related}
While the research in classification of taxiway line markings is still in its preliminary stages, the domain of car lane detection has seen substantial advancements in road labeling techniques. The knowledge derived from detecting car lanes not only provides fundamental perspectives that guide our comprehension of detecting taxiway markings but also highlights research gaps in car lane labeling methods which can be avoided and guide innovative approaches for taxiway marking detection. Therefore, this section begins by presenting notable works from the car lane detection domain and highlights their research gaps.

Some researchers have used classification techniques for lane detection, where images are segmented into grids for row-based lane location identification  \cite{qin2020ultra,yoo2020end,chougule2018reliable}. However, these methods lack precision and might miss some lanes. 
To ensure consistent lane detection, techniques such as parametric curve modeling \cite{proesmans2018towards,wang2020polynomial,tabelini2021polylanenet,liu2021end,feng2022rethinking} and keypoint association \cite{qu2021focus,wang2022keypoint,xu2022rclane} have been used. Even though these methods acquire high results, they can struggle in adverse weather conditions.
Andrei \etal \cite{andrei2022} have utilized probabilistic Hough transform and dynamic parallelogram region of interest (ROI) to develop a lane detection system. It was implemented on video sequences with manual ROI definition but encountered challenges with capturing curved line endpoints, which are crucial for complete lane boundary delineation in real-world scenarios. Chen \etal \cite{Chen2015} detected road markings using machine learning with binarized normed gradient detection and principal component analysis (PCA) classification, achieving an accuracy of 96.8\%. However, the dependency on PCA limited the adaptability for dynamic scenarios and challenging environments. Similarly, the method proposed by Ding \etal \cite{ding2020}, which combined PCA and support vector machine (SVM), achieved a detection accuracy of 94.77\%. However, it struggled to detect road markings due to the failure of the ROI extraction, in the case of reflection, occlusion and shadow.

Gupta \etal \cite{gupta2018} introduced a real-time framework for camera-based road and lane markings, using techniques such as spatio-temporal incremental clustering, curve fitting, and Grassmann manifold learning. The real-time nature of this approach brings challenges in processing efficiency and speed, especially with increasing dataset complexities. Jiao \etal \cite{Jiao2019} proposed an adaptive lane identification system using the scan line method, lane-voted vanishing point, and multi-lane tracking Kalman filtering,  achieving a 93.4\% F1-Score. This approach lacked the adaptability to diverse conditions across different datasets and real-world scenarios.

Kiske \cite{kiskeautomatic} developed an autonomous labeling system for identifying highway lane markings using Velodyne lidar, HDR video streams, and high-precision GPS. However, the reliance on multiple data sources makes it less cost-effective and more complex for wide-scale implementations. Muthalagu \etal \cite{muthalagu2020} proposed a vision-based algorithm for lane detection in self-driving cars, which used polynomial regression, histogram analysis, and a sliding window mechanism for detecting both straight and curved lines. The algorithm achieved notable accuracy, but its high computational demand poses scalability issues, and its limitation in detecting steep foreground curves presents challenges in dynamic terrains.

\begin{figure*}[h!tbp]
    \centering
    \includegraphics[width=\linewidth]
    {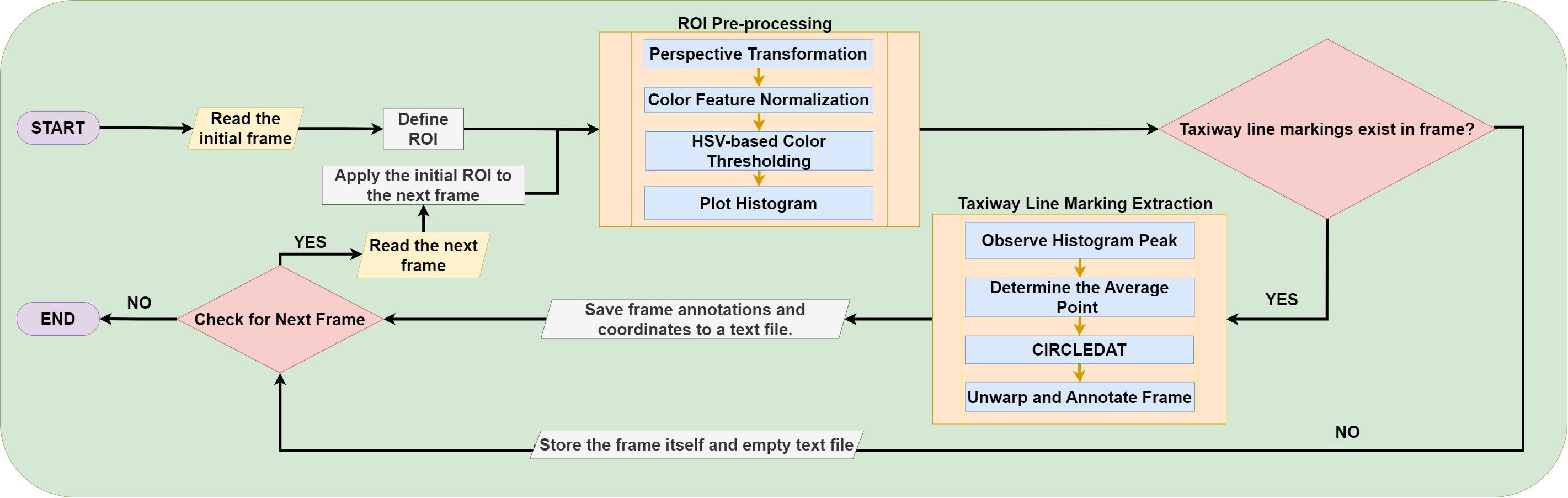}
    \caption{Framework of ALINA}
    \label{fig:label_1}
\end{figure*}
Contour-based detection and line extraction method (CDLEM) was initially used for labeling the taxiway line markings from AssistTaxi dataset \cite{ganeriwala2023assisttaxi}. The Canny edge detection identified the line marking's edges. Subsequently, the Hough transform, Ramer-Douglas-Peucker, and Bresenham's algorithms were utilized to identify and label both straight and curved taxiway line markings. The limitation this approach posed was the requirement to outline an ROI around taxiway line marking for every scenario shift.

While labeling car lanes has provided a guiding example for an end to end lane detection system \cite{Zhang2018end,alylanquest2015,Neven2018towards}, a direct comparison of lane detection to labeling taxiway line markings is not being made in this research. The distinct semantics and attributes associated with the car lane datasets and airport taxiways dataset pose different set of challenges. For example, the airport taxiway dataset has different layouts, diverse markings, and presence of aircraft on airport taxiways. Transfer learning can be one of the approaches to address the differences and identify the commonalities within the two datasets \cite{ganeriwalacross2023}, however it requires the taxiways to be extensively labeled. Therefore, our research emphasizes creating and evaluating algorithms for labeling specific to airport taxiways, rather than contrasting them with car lane datasets.

\section{Methodology}
\label{method}

In this section, we provide a detailed overview of the ALINA framework using \cref{fig:label_1} (each subsection elaborates elements from the figure), and we discuss its steps referencing image instances from \cref{fig:roi_preprocessing} and \cref{fig:marking_extraction}.
\subsection{Frame Representation}

While reading a frame, ALINA stores each x, y coordinate and its red, green, blue (RGB) channel values from the frame into a multi-dimensional array, as represented in \cref{eq:1}. 
\begin{equation} \label{eq:1}
    A[i, j, k] = I[i, j, k]
\end{equation}
where, i, j, and k denote the row, column, and channel indices of the frame, respectively. This formula copies the pixel values of the frame at location (i, j) in the \(k^{\text{th}}\) color channel into the corresponding location in the array. 

\subsection{Interactive ROI Definition on the Initial Frame}


The user provides source points to ALINA for creating a trapezoidal region of interest (ROI) on the initial frame of the video, as shown in \cref{fig:roi_preprocessing} (a), (e), (f). It is drawn in the vicinity of where the camera expects to see the taxiway's line markings. ALINA treats the ROI as a constraint to limit the search area for taxiway's line markings, resulting in reduced computational complexity and improved detection accuracy \cite{hoang2019}. 
The source points, which correspond to the vertices of ROI, are denoted as follows: \([x_1, y_1],\ [x_2, y_2],\ [x_3, y_3],\ [x_4, y_4]\).

\subsection{Perspective Transformation of the ROI}


The next step in ALINA is warping the perspective. The trapezoidal ROI is warped into a bird's eye view.  \cref{fig:roi_preprocessing} (b), (f), (j) illustrates this transformation. The destination points, which correspond to the vertices of a rectangle defining the bird's eye view, are represented as: \([x_1', y_1'],\ [x_2', y_2'],\ [x_3', y_3'],\ [x_4', y_4']\).


It enables a more comprehensive and consistent view of the taxiway and its features, allowing for improved detection of line markings irrespective of their orientation or curvature \cite{al2023}. 

Given the pairs of source and destination points, a matrix M is derived by solving a system of linear equations formed from the pixel correspondences. The matrix essentially maps any pixel in the trapezoidal ROI to its corresponding pixel position in the bird's eye view.

The matrix M is represented in \cref{eq:93}:
\begin{equation} \label{eq:93}
M = \begin{bmatrix}
m_{11} & m_{12} & m_{13} \\
m_{21} & m_{22} & m_{23} \\
m_{31} & m_{32} & 1
\end{bmatrix}
\end{equation}
with each \(m_{ij}\) computed based on the defined system of equations.

Once M is computed, it's applied to each pixel in the trapezoidal ROI to achieve its position in the bird's eye view. This transformation is attained using \cref{eq:94}:
\begin{equation} \label{eq:94}
    \text{t}(x, y) = s\left(\frac{M_{11}x + M_{12}y + M_{13}}{M_{31}x + M_{32}y + M_{33}}, \frac{M_{21}x + M_{22}y + M_{23}}{M_{31}x + M_{32}y + M_{33}}\right)
\end{equation}

Where \(t(x,y)\) and \(s(x,y)\) denote pixel coordinates in the bird's eye view and trapezoidal ROI respectively.

\subsection{Color Feature Normalization}

To accurately distinguish line markings within the ROI, ALINA performs normalization of color characteristics. The normalization phase consists of the following key steps:

\begin{enumerate}
    \item \textbf{\(RGB\) to \(HSV\) Conversion:} The \(RGB\) color space of ROI is transformed to the hue, saturation, and value \((HSV)\) color space. Given a pixel's \(RGB\) values, denoted by \((R, G, B)\), the transformation into the \(HSV\) space is depicted as \((R, G, B) \rightarrow (H, S, V)\). This transformation is performed because \(HSV\) factors in variations induced by lighting conditions and intrinsic color properties, providing a robust representation of color in an image \cite{shuhua2010}.
    \item \textbf{Component Decomposition:} The \(HSV\) is decomposed into its individual channels: \(H, S, V\).
    \item \textbf{Normalization:} Every component \(X\), where \(X \in \{H, S, V\}\), undergoes 8-bit min-max normalization as shown in \cref{eq:normalization}: 
    \begin{equation}
    X_{\text{norm}} = \frac{X - \min(X)}{\max(X) - \min(X)} \times 255
    \label{eq:normalization}
    \end{equation}
    where \(\min(X)\) and \(\max(X)\) denote the minimum and maximum values of the component \(X\), respectively. This ensures the values are scaled to fit within the \([0, 255]\) range.
    \item \textbf{Component Recombination:} The normalized components, denoted as \(H_{\text{norm}}, S_{\text{norm}}, V_{\text{norm}}\), are recombined to form the normalized \(HSV\) ROI, represented as \(HSV_{\text{norm}}\).
\end{enumerate}

Following this normalization process, the color features of the frame are standardized, as illustrated in \cref{fig:roi_preprocessing} (c), (g), (k). This ensures uniform contribution from each feature to the final representation, facilitating precise color thresholding for taxiway line marking detection in the subsequent step.


\subsection{HSV-based Color Thresholding}

Color thresholding is an essential technique in image segmentation, leveraging color information to partition image pixels into meaningful regions \cite{cheng2001}. In the context of ALINA, this technique finds its application in isolating the taxiway line markings from the rest of the regions within the ROI, using the \(HSV\) color space. Hue captures the wavelength of color, while saturation measures the intensity, and value quantifies the brightness. 

Determining the precise range for \(HSV\) that corresponds to the taxiway line markings necessitated a series of empirical tests. Multiple frame samples were analyzed and a frequency distribution of \(HSV\) values for taxiway line marking regions were plotted. Peaks in this distribution, indicative of dominant \(HSV\) values for taxiway line markings, helped in ascertaining the lower and upper bounds of \(H, S, V\), i.e. (0, 70, 170) and (255, 255, 255), respectively.

For a pixel p with \(HSV\) values denoted by \(H(p)\), \(S(p)\), \(V(p)\), the color thresholding function \(\Theta(p)\)is defined in \cref{eq:color_thresholding}: 

\begin{equation}
\Theta(p) = 
\begin{cases} 
255 & \text{if } 0 \leq H(p) \leq 255 \\
& \text{and } 70 \leq S(p) \leq 255 \\
& \text{and } 170 \leq V(p) \leq 255 \\
0 & \text{otherwise}
\end{cases}
\label{eq:color_thresholding}
\end{equation}

\begin{figure}[t]
    \centering
    \includegraphics[width=\linewidth,height=6cm]{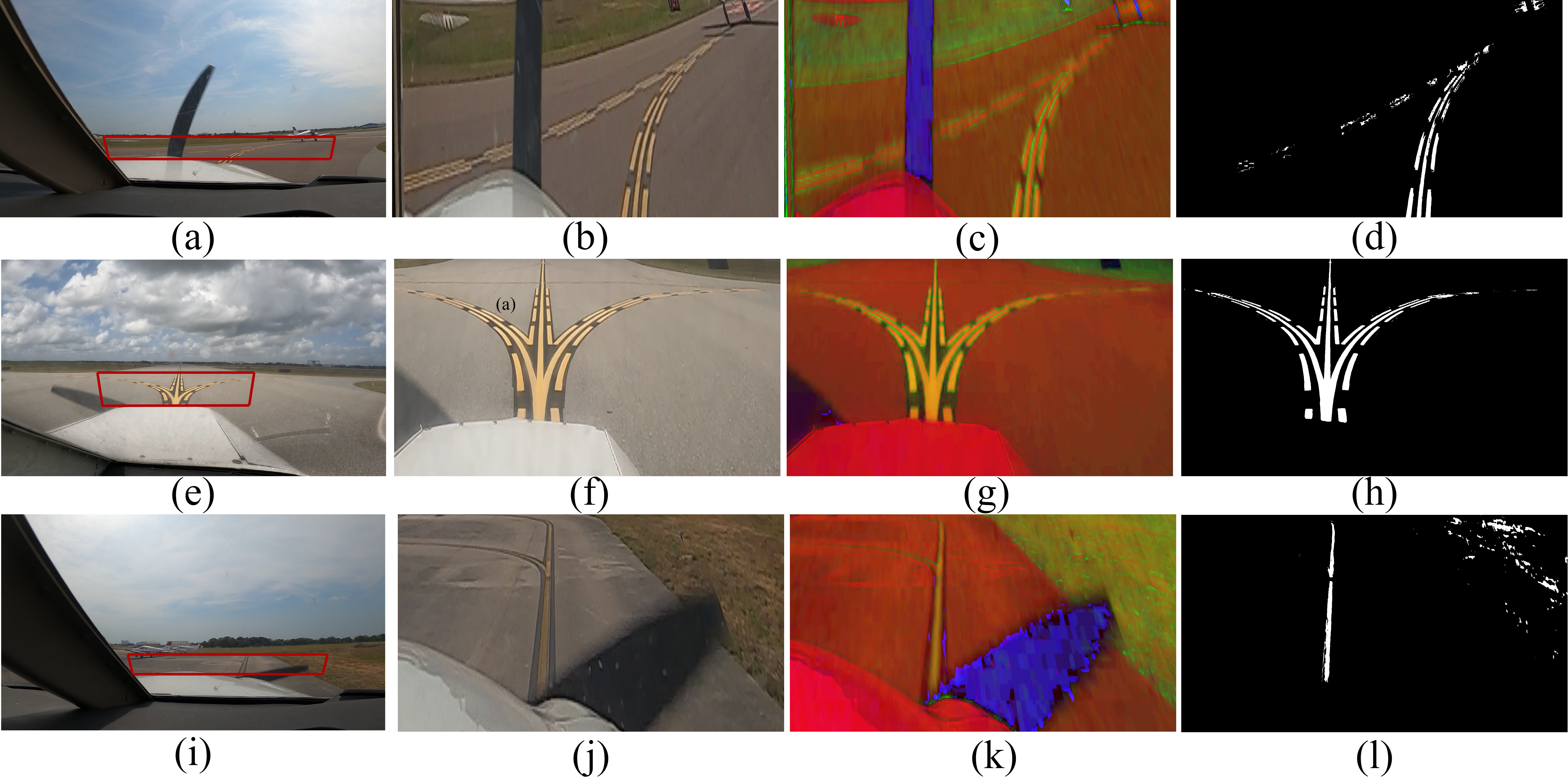}
    \caption{(a),(e),(i): ROI Definition, (b),(f),(j): Perspective Transformation, (c),(g),(k): Color Feature Normalization, (d),(h),(l): HSV-based Color Thresholding}
    \label{fig:roi_preprocessing}
\end{figure}

Here, \(\Theta(p)\) produces a binary outcome for each pixel in the ROI: a value of 255 (white) indicates a pixel belonging to the taxiway line marking, while 0 (black) denotes a pixel that does not. However, after applying HSV-based color thresholding, some non-taxiway line marking pixels within the ROI still matched the specified range, as shown in \cref{fig:roi_preprocessing} (d), (h), (l). ALINA addresses this in the subsequent step.


\subsection{Histogram-based Analysis of the Thresholded ROI}

The histogram analysis focuses on the vertical projection of white pixels in the thresholded ROI, analyzing the spatial distribution and density of the taxiway line markings.

For a thresholded ROI with dimensions \(W \times H\), where \(W\) represents the width (number of columns) and \(H\) represents the height (number of rows), the binary representation of a pixel at position \(i, j\) is defined in the \cref{eq:222}, where \(i\) is the column index and \(j\) is the row index:

\begin{equation} \label{eq:222}
B(i,j) = 
\begin{cases} 
1 & \text{if pixel at }(i,j) \text{ is white} \\
0 & \text{otherwise}
\end{cases}
\end{equation}

The vertical histogram, denoted as \(H_{\text{vert}}(i)\), quantifies the column-wise distribution of white pixels and is calculated using \cref{eq:223}:

\begin{equation} \label{eq:223}
H_{\text{vert}}(i) = \sum_{j=0}^{H} B(i, j)
\end{equation}

For each column \(i\), this equation aggregates the presence of white pixels across all rows \(j\), offering a count of white pixels for that column. When plotted against the column index \(i\), the histogram produced accentuates the density of white pixels along the y-axis. Peaks in this histogram, denote the presence and spatial positioning of taxiway line markings within the ROI.


\subsection{Identifying Line Markings and Mitigating False Detections}

The histogram peak value, represented by \(H_{p}\), indicates the highest density of white pixels in a particular column of the histogram. It was important to identify whether a peak in the histogram truly represents a taxiway line marking or in contrast, is a result of noise or other disturbances. We identified the presence of a taxiway line marking in the ROI using a threshold value. 

\begin{figure}[t]
    \centering
    \includegraphics[width=\linewidth, height=6cm]{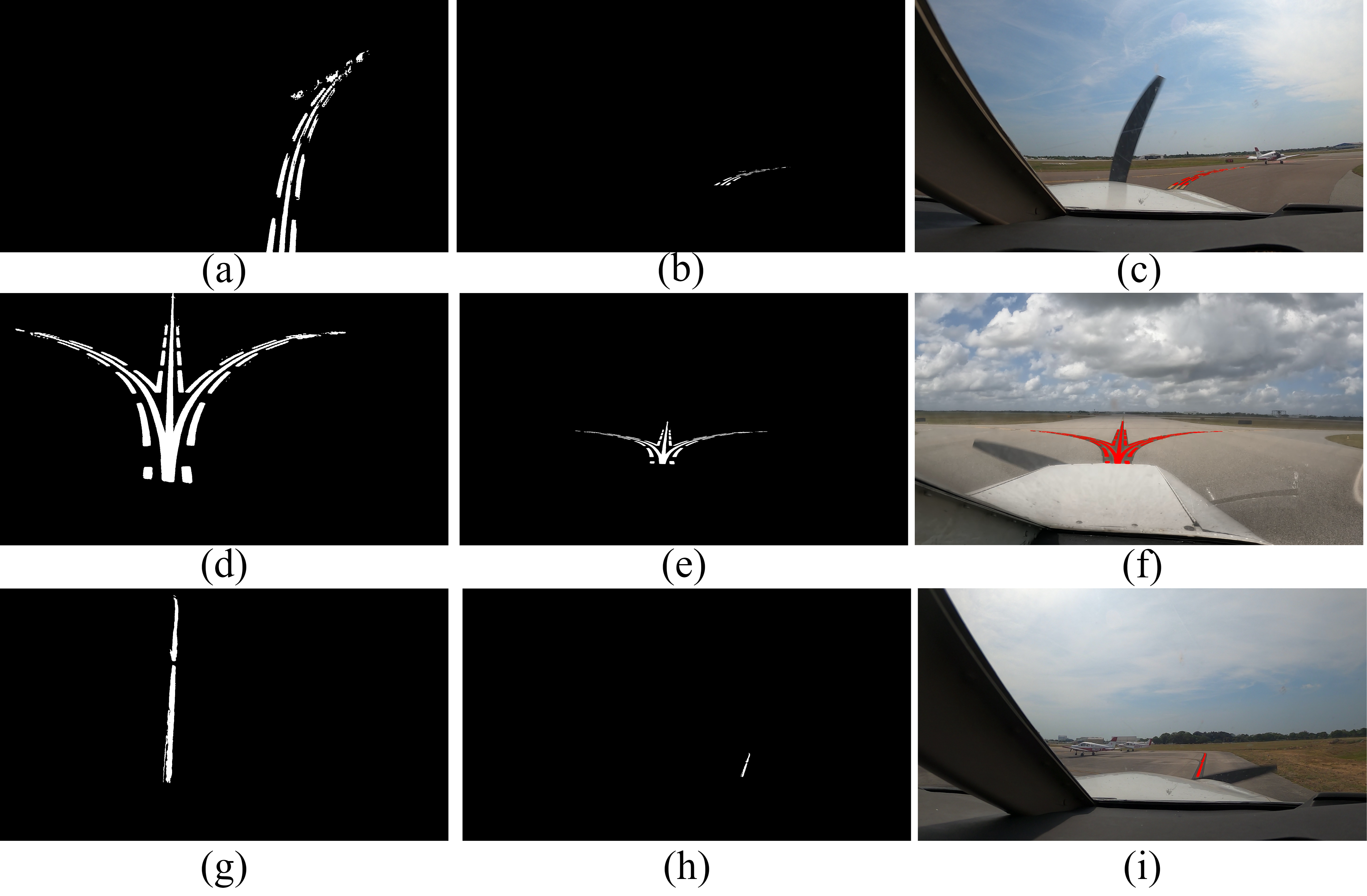}
    \caption{(a),(d),(g): CIRCLEDAT, (b),(e),(h): Frame Unwarping, (c),(f),(i): Taxiway Line Marking Annotation}
    \label{fig:marking_extraction}
\end{figure}

\subsection{CIRCLEDAT: Circular Threshold Pixel Discovery and Traversal Algorithm}

The \textsc{CIRCLEDAT} algorithm assists in isolating the pixels that correspond to the taxiway line markings, irrespective of their dimension or curvature, and eliminate all other pixels from the ROI. For an ROI \( \mathcal{I} \) with dimensions \( W \times H \), an initial coordinate pair \( (x, y) \), a radius \( \theta \), a set \( \mathcal{V} \) for recording visited pixels, and an array \( \mathcal{L} \) for collecting taxiway line marking pixels, the algorithm executes as follows:

\begin{enumerate}
    \item \textbf{Initialization:} The starting point \( (x, y) \) is pushed onto a stack \( S \) and recorded in \( \mathcal{V} \). A set \( \mathcal{R} \) is created containing all pixel coordinates within the circular distance.

    \item \textbf{Exploration:} While \( S \) is not empty, the top coordinate \( (x, y) \) is popped. If \( \mathcal{I}[y][x] = 255 \), which indicates a white pixel, \( (x, y) \) is added to \( \mathcal{L} \). For each offset \( (i, j) \) in \( \mathcal{R} \):
    \begin{itemize}
        \item Calculate new coordinates \( (x_{\text{new}}, y_{\text{new}}) = (x+i, y+j) \).
        \item If \( (x_{\text{new}}, y_{\text{new}}) \in \mathcal{V} \) or \( (x_{\text{new}}, y_{\text{new}}) \) is outside \( 0 \leq x_{\text{new}} < W \) and \( 0 \leq y_{\text{new}} < H \), continue to the next offset.
        \item Otherwise, push \( (x_{\text{new}}, y_{\text{new}}) \) onto \( S \) and record in \( \mathcal{V} \).
    \end{itemize}

    \item \textbf{Result:} Once \( S \) is exhausted, the algorithm returns \( \mathcal{L} \), representing all the white pixels, which correspond to the taxiway line markings.
\end{enumerate}

The \cref{fig:marking_extraction} (a), (d), (g) represent the output generated after applying the CIRCLEDAT algorithm (\ref{alg:threshold}) within the ROI. Through a depth-first search approach, \textsc{CIRCLEDAT} ensures a comprehensive exploration of pixels pertaining to the taxiway line markings. The \( \mathcal{V} \) set aids in efficient traversal by preventing revisits, and the use of \( \mathcal{R} \) ensures a circular-based neighborhood exploration. 




\begin{algorithm}[h]
\caption{Circular Threshold Pixel Discovery and Traversal}
\label{alg:threshold}
\begin{algorithmic}[1]
\Function{CIRCLEDAT}{$image$}
    \State $stack \gets [(x, y)]$
    \State $visited.add((x, y))$
    \State $circular\_range \gets list(product(range(-\theta, \theta+1), repeat \gets 2))$
    \State $circular\_range.remove((0, 0))$
    \While{$stack$ is not empty}
        \State $x, y \gets stack.pop()$
        \If{$image[y][x] = 255$}
            \State $add\ (x, y)\ to\ line\_marking\_pixels$
            \For{$i, j \gets circular\_range$}
                \State $x\_new \gets x + i$
                \State $y\_new \gets y + j$

                \If{$(x\_new, y\_new) = visited$}
                    \State \textbf{continue}
                \EndIf

                \If{$x\_new, y\_new < 0\ or\ x\_new, y\_new \geq image.width, image.height$}
                    \State \textbf{continue}
                \EndIf

                \State $add\ (x\_new, y\_new)\ to\ stack$
                \State $visited.add((x\_new, y\_new))$
            \EndFor
        \EndIf
    \EndWhile
    \State $return\ line\_marking\_pixels$
\EndFunction
\end{algorithmic}
\end{algorithm}

\subsection{Frame Unwarping and Annotating the Taxiway Line Marking Pixels}


The final stage of labeling the frame involves an inverse perspective transformation that returns the warped ROI to its original view, as shown in \cref{fig:marking_extraction} (b) ,(e), (h).
After obtaining the unwarped ROI, white pixels are located and mapped onto the original frame using color red to clearly identify the taxiway line markings, as illustrated in \cref{fig:marking_extraction} (c), (f), (i).
This representation provides a precise and clear depiction of the taxiway line markings in the original frame. In addition, a text file is generated containing the x, y coordinates of all pixels corresponding to the taxiway line markings.

Having completed the labeling of the video's initial frame, ALINA proceeds to label the subsequent frames using the same process and the ROI established for the first frame, as illustrated in \cref{fig:label_1}.

\section{Experimental Results}
\label{results}

\subsection{ALINA Performance: Specifications and Scenarios}

The detailed processing time breakdown for ALINA when labeling a frame is presented in \cref{tab:processing-time}. On average, ALINA requires 50.9 milliseconds (ms) to label a single frame, yielding a rate of approximately 19.65 frames per second (fps). In the course of our systematic labeling process, ALINA effectively labeled 60,249 frames on a Linux system equipped with an Intel Core i7-9700K CPU clocked at 3.60GHz and 15GB RAM, operating on Ubuntu 22.04.1 LTS. The algorithm was developed in the PyCharm IDE with Python 3.8.15, harnessing essential libraries such as OpenCV, NumPy, MatPlotLib, and statistics.

\begin{table}[htbp]
  \caption{Processing time analysis of ALINA (in Milliseconds).}
  \centering
  \begin{tabular}{@{}lr@{}}
    \toprule
    \textbf{Process} & \textbf{Time (ms)} \\
    \midrule
    Perspective Transformation & 4.41 \\
    Color Feature Normalization & 5.91 \\
    HSV-based Color Thresholding & 1.05 \\
    Histogram Analysis & 28.71 \\
    CIRCLEDAT & 3.33 \\
    Projection Remapping & 6.68 \\
    \midrule
    \textbf{Total} & \textbf{50.09} \\
    \bottomrule
  \end{tabular}
  \label{tab:processing-time}
\end{table}

As mentioned earlier, we applied ALINA on a subset of the AssistTaxi dataset, consisting of 60,249 frames extracted from three distinct videos, each with unique camera angles. \cref{fig:roi_preprocessing} and \cref{fig:marking_extraction} demonstrates ALINA's consistent labeling performance across these frames, unaffected by differing camera angles or weather conditions. Specifically, in \cref{fig:roi_preprocessing} (a) from the first video, ALINA labeled the taxiway line marking between two aircrafts. In \cref{fig:roi_preprocessing} (e) from the second video, it labeled three directional taxiway line markings: left, straight, and right. Lastly, in \cref{fig:roi_preprocessing} (i) from the third video, ALINA labeled the taxiway line marking in front of the aircraft, which is situated between stationary aircraft on the left and grassy area on the right.

\subsection{Ablation Study on Threshold Value}

Initially, the decision was kept binary, in which there was no set threshold, as shown in \cref{eq:224}. This approach considered even a slight peak in the histogram as a taxiway line marking, leading to a high percentage of false positives, as shown in \cref{tab:fp_comparison}. 

\begin{equation} \label{eq:224}
\Delta(p) = 
\begin{cases} 
1 & \text{if } H_p > 0 \\
0 & \text{otherwise}
\end{cases}
\end{equation}

Here \(\Delta(p)\) indicates the presence (1) or absence (0) of the taxiway line marking.

Through empirical testing and histogram analyses, we identified that true taxiway line markings consistently yielded peak values significantly above sporadic noise. As shown in \cref{tab:fp_comparison}, setting a threshold at 75 led to a substantial fall in the percentage of false positives, but by raising the threshold to 150, the false positives were dropped to 0. Hence, the \( T_{optimal} \) was established at 150, using the equations \ref{eq:225} and \ref{eq:226}. This was primarily because true taxiway line markings consistently resulted in columns extending well over 150 white pixels, while disturbances never reached this level. 

\begin{table}[htbp]
  \caption{Threshold Value Comparison}
  \centering
  \begin{tabular}{@{}lc@{}}
    \toprule
    \textbf{Threshold Value} & \textbf{False Positives (\%)} \\
    \midrule
    0 & 83.33 \\
    75 & 22.22 \\
    150 & 0 \\
    \bottomrule
  \end{tabular}
  \label{tab:fp_comparison}
\end{table}
\begin{equation} \label{eq:225}
T_{\text{optimal}} = \underset{T}{\mathrm{argmin}} \left( F_{\text{false positive}}(T) - F_{\text{true positive}}(T) \right)
\end{equation}
\begin{equation} \label{eq:226}
\Delta(p) = 
\begin{cases} 
1 & \text{if } H_p >= T_{optimal} \\
0 & \text{otherwise}
\end{cases}
\end{equation}

Once the taxiway line marking is detected using \(\Delta_{150}(p)\), ALINA extracts its centroid coordinate, initializing the subsequent CIRCLEDAT algorithm. In contrast, if there is no taxiway line marking in the ROI, ALINA simply stores the frame along with an empty text file, signifying that there is no taxiway line marking present in the frame, as shown in \cref{fig:label_1}.

\subsection{Generating a Context-based Edge Map Set}

To assess ALINA and CDLEM's effectiveness, we manually created a set of context-based edge maps (CBEM), which emphasize the edge pixel presence, edge corner localization, thick edge occurrence, and edge connectivity \cite{canny1986computational}. We prioritized the detection of the edges of a taxiway line marking as our primary validation metric, because this alone provides precise information into the line marking's position within the frame. Our approach for developing the CBEM was as follows:
\begin{enumerate}
    \item \textbf{Outlining the Contour Region:} The taxiway line markings in the frames were outlined manually to create an accurate reference without including any other edge details from the frame.
    \item \textbf{Pre-processing:} The contour region was transformed to grayscale and then subjected to Gaussian blur, which helped in reducing the visual noise and improved the clarity of edges. Consequently, the gradient amplitude and direction were calculated and non-maximum suppression was applied to eliminate the non-edge pixels.

    \item \textbf{Edge Detection with the Canny Algorithm:} We used the Canny algorithm for precise edge extraction of taxiway line markings. We also employed an automated method for selecting the upper and lower thresholds \cite{Rosebrock2015}. This method computes the median pixel intensity \(v\) of an image and subsequently determines the thresholds using equations \ref{eq:338} and \ref{eq:339}:
    \begin{equation} \label{eq:338}
        \text{lower} = \max(0, (1.0 - \sigma) \times v)
    \end{equation}
    \begin{equation} \label{eq:339}
        \text{upper} = \min(255, (1.0 + \sigma) \times v)
    \end{equation}
    where \( \sigma \) is a coefficient, defaulting to 0.33, for refining the thresholds.
\end{enumerate}

\begin{figure} 
    \centering
    \includegraphics[width=\linewidth]
    {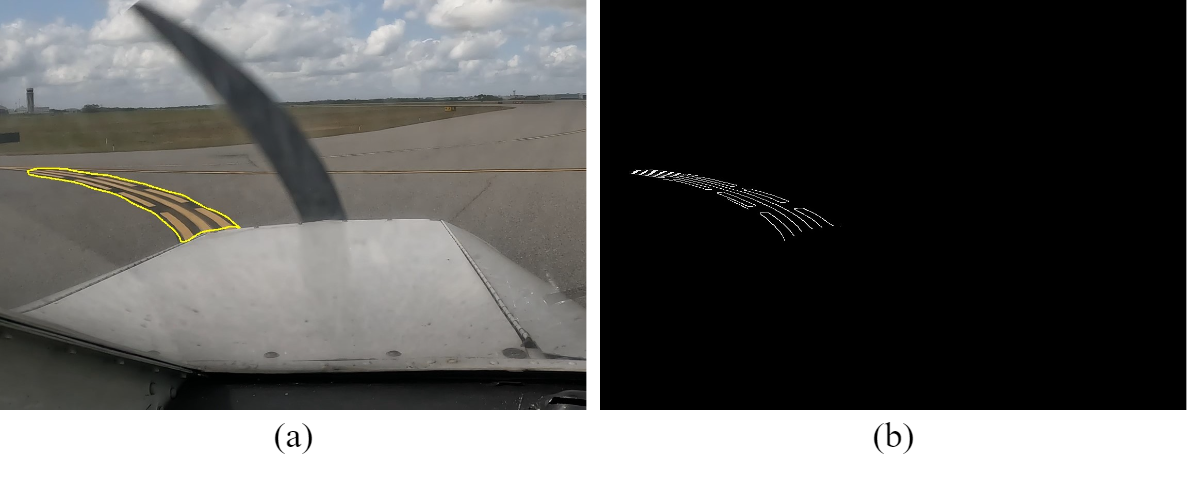}
    \caption{(a) Manually Outlined Contour Region (b) CBEM}
    \label{fig:label_52}
\end{figure}

The process of manually outlining the contour region around the taxiway line marking and generating CBEM for a frame is illustrated in \cref{fig:label_52} (a) and (b) respectively. The \cref{fig:main} (e) shows the output of ALINA on the same frame.

From the 60,249 frames, we selected a set of 120 frames to construct the CBEM's. The 60,249 frames spanned from three videos with durations of 11.47, 1.34, and 3.23 minutes. Our objective was to identify the frames representing scenario shifts. Instead of conducting a granular frame-by-frame analysis, we reviewed the videos comprehensively and marked the specific frames that captured the scenario shifts, amounting to a total of 120 images.

Our selection is underpinned by the Law of Large Numbers (LLN), as illustrated in \cref{eq:333}:
\begin{equation} \label{eq:333}
\bar{X}_n \rightarrow \mu \quad \text{as} \quad n \rightarrow \infty
\end{equation}
where $\bar{X}_n$ represents the sample mean and $\mu$ is the expected population mean \cite{feller1991introduction}. This indicates that our subset, if representatively selected, provides a reliable approximation of the comprehensive dataset's attributes.

Furthermore, the Central Limit Theorem (CLT) \cite{feller1991introduction} also reinfornces our approach, as expressed in \cref{eq:334}:
\begin{equation} \label{eq:334}
 \frac{S_n - n\mu}{\sigma \sqrt{n}} \rightarrow N(0,1)
\end{equation}
where $S_n = X_1 + X_2 + \ldots + X_n$ and \( X_i \) are independent, identically distributed random variables, \( n \) is the sample size, \( \mu \) is the population mean, and \( \sigma \) is the standard deviation. The CLT's cornerstone assertion, relevant in our context, is that with a sufficiently extensive sample size, the sample mean's distribution gravitates towards a normal distribution. This holds irrespective of the originating population's distribution. Therefore, the mean distribution extrapolated from all possible 120-frame subsets is poised to achieve normality.
Leveraging the Central Limit Theorem, our diverse 120-frame sample is statistically representative of the entire 60,249-frame dataset. With the backing of both LLN and CLT, our method ensures a robust evaluation of ALINA and CDLEM using CBEM set.




\subsection{Sliding Window Vs CIRCLEDAT}

In the study by Muthalagu et al. \cite{muthalagu2020}, a sliding window (SW) search algorithm detected lane line marking pixels with a time complexity of $O(m \times n)$, where $m$ and $n$ represent the height and width of the frame, respectively. In contrast, our work introduces the innovative CIRCLEDAT algorithm, which pinpoints line marking pixels with a significantly reduced time complexity of $O(k)$, where k is number of pixels corresponding to the line marking in a given frame. Prior to introducing CIRCLEDAT, we used the SW search algorithm to detect taxiway line marking pixels. \cref{tab:comparison} compares the performance of both algorithms when tested on the taxiway dataset frames.
\begin{table}[htbp]
  \caption{Comparative Analysis of Pixel Detection Algorithms}
  \centering
  \begin{tabular}{@{}lcc@{}}
    \toprule
    \textbf{Algorithm} & \textbf{Time Complexity} & \textbf{Processing Time (ms)} \\
    \midrule
    SW Search & \(O(m \times n)\) & 10.90 \\
    CIRCLEDAT & \(O(k)\) & 3.33 \\
    \bottomrule
  \end{tabular}
  \label{tab:comparison}
\end{table}

\subsection{Performance Evaluations}

We evaluated ALINA's performance against CDLEM using CBEM set. By comparing x and y coordinates from both the CBEM and ALINA or CDLEM, we calculated true positives (TP) and false negatives (FN). TP indicates accurate identification of taxiway line marking pixels, while FN denotes missed pixels that should have been identified as part of the taxiway line marking.

The recall or detection rate, essential for evaluating object detection algorithms, gauges an algorithm's accuracy in identifying the particular objects in a frame \cite{padilla2021comparative} \cite{diebold2002comparing}. In airport taxiway line marking detection, missing a marking can pose safety risks, underscoring the importance of comprehensive identification. The detection rate is calculated as the ratio of TP values to the sum of TP and FN values, as shown in \cref{eq:211}. 

\begin{equation} \label{eq:211}
    Detection\ Rate\ (Recall) =\ \frac{TP}{TP + FN}
\end{equation}

ALINA and CDLEM achieved detection rates of 98.45\% and 91.14\%, respectively, as shown in \cref{table:label_81}. Additionally, in terms of processing time, ALINA processed a frame in 50.09 ms, corresponding to approximately 19.65 fps, whereas CDLEM took 120.35 ms per frame, translating to roughly 8.30 fps.

ALINA's superior performance is attributed to several key features. Its perspective transformation provides a bird's-eye view of taxiway, eliminating distortions and offering clarity in distinguishing line markings from anomalies—a challenge for CDLEM due to varying distances and angles. Unlike CDLEM, which may miss essential pixels using the Hough Transform and curve fitting, ALINA's shift to the \(HSV\) color space, prioritizing H, S, and V components, enhances taxiway line marking detection even under varying weather. The CIRCLEDAT algorithm within ALINA swiftly captures all crucial pixels of taxiway markings, regardless of their shape or fragmentation. A notable limitation of CDLEM is its need to define a new ROI for each scenario shift, resulting in 120 scenarios for 60,249 frames, demanding extensive video pre-viewing. In contrast, ALINA only requires an ROI for the initial frame of a video, applied consistently to all following frames, leading to just 3 ROIs for the same number of frames.

\begin{table}[htbp]
\caption{Comparing ALINA Vs CDLEM}
\label{table:label_81}
\centering
\begin{tabular}{@{}lcc@{}}
  \hline
  \textbf{Algorithm} & \textbf{Detection Rate (\%)} & \textbf{Processing Time (ms)} \\
  \hline
  ALINA & 98.45 & 50.09 \\
  CDLEM & 91.14 & 120.35 \\
  \hline
\end{tabular}
\end{table}
\section{Conclusion}
\label{conclude}
In this work, we propose ALINA, a novel annotation framework, primarily designed for labeling pixel coordinates in taxiway datasets. This approach streamlines the annotation process, significantly reducing cost and manual labor needed for precise labeling in these contexts. We also propose a traversal algorithm CIRCLEDAT, which determines the pixels corresponding to the taxiway line markings. We provide a comparative analysis with the sliding window search algorithm and evaluate the performance of the framework on a subset of the AssistTaxi dataset. We have tested ALINA with labels generated for 60,249 frames, and evaluated it with a context-based edge map (CBEM) set which was generated manually. We also provide theoretical analysis and a comparative study for ALINA to Contour-Based Detection and Line Extraction Method (CDLEM). In the future, we aim to evaluate ALINA for annotating car lane datasets, with the CIRCLEDAT algorithm being utilized to identify pixel coordinates of road lane markings.
{
    \small
    \bibliographystyle{ieeenat_fullname}
    \bibliography{main}
}


\end{document}